# BEKG: A Built Environment Knowledge Graph

Xiaojun Yang[a], Haoyu Zhong[a], Penglin Du[a], Keyi Zhou[a], Xingjin Lai[a], Zhengdong Wang[a], Yik Lun Lau[b], Yangqiu Song[b], Liyaning Tang[c]*

[a] *School of Information Engineering, Guangdong University of Technology, Guangzhou, China*

[b] *Department of Computer Science and Engineering, The Hong Kong University of Science and Technology, Hong Kong, China*

[c] *School of Architecture and Built Environment, The University of Newcastle, New South Wales, Australia*

*Corresponding author at: School of Architecture and Built Environment, The University of Newcastle, New South Wales, Australia

Email address at: maggie.tang@newcastle.edu.au (L. Tang)

## Abstract

Practices in the built environment have become more digitalized with the rapid development of modern design and construction technologies. However, the requirement of practitioners or scholars to gather complicated professional knowledge in the built environment has not been satisfied yet. In this paper, more than 80,000 paper abstracts in the built environment field were obtained to build a knowledge graph, a knowledge base storing entities and their connective relations in a graph-structured data model. To ensure the retrieval accuracy of the entities and relations in the knowledge graph, two well-annotated datasets have been created, containing 2,000 instances and 1,450 instances each in 29 relations for the named entity recognition task and relation extraction task respectively. These two tasks were solved by two BERT-based models trained on the proposed dataset. Both models attained an accuracy above 85% on these two tasks. More than 200,000 high-quality relations and entities were obtained using these models to extract all abstract data. Finally, this knowledge graph is presented as a self-developed visualization system to reveal relations between various entities in the domain. Both the source code and the annotated dataset can be found here: https://github.com/HKUST-KnowComp/BEKG.

## 1   Introduction

Data has been playing a more important role than ever when it comes to utilizing productivity and strategic resources to drive economic growth and social progress nowadays. Existing studies have suggested a wide range of possible artificial intelligence applications in various data analysis tasks in the field of the built environment. To illustrate, Wei et al. (2018) leveraged machine learning models to analyze electricity consumption behavior at the city level. Their approach successfully yielded a more accurate prediction on the city level and had the advantage of implementing a more responsive and scalable demand-side management system. Yu et al. (2010) proposed the decision tree-based models to forecast building energy demand levels for a more intuitive forecasting method instead. Newton (2018) developed a multi-objective optimization algorithm to classify 3D building features with 3D convolutional neural networks. Ghadai et al. (2018) proposed a new feature recognition framework that uses deep learning to foresee challenges in borehole fabrication. There is a noticeable trend in embracing more advanced artificial intelligence tools to conduct analysis and forecasting in the field.

Besides recent improvements in data analysis and forecasting tools in the area, data collection and knowledge management are also essential to support decision-making and managerial decisions for future built environment projects. Since projects in the field are of interdisciplinary nature, it requires expertise across multiple domains to understand issues thoroughly and create strategic plans. Despite knowledge management being a challenging problem, it is valuable to assist project managers in making informed decisions while knowledge is reusable in other construction projects. For example, knowledge of construction projects in the past can be helpful for managers in assessing project risks in the future (Tah and Carr, 2001; Nieto-Morote and Ruz-Vila., 2011). Also, the distilled wisdom from the past construction projects could better assist managers in analyzing the construction project cooperation network (Sun et al., 2019). As the amount of knowledge generated from projects increases, knowledge management becomes a critical problem for practitioners in handling project details precisely and efficiently. Therefore, Udeaja et al. (2008) developed a prototype to capture and reuse the knowledge harvested from construction projects. Ozorhon et al. (2013) proposed a web-based knowledge management tool to create organizational memory for capturing, storing, sharing, and using project and corporate information to manage the normative knowledge.

Apart from the knowledge management mentioned above, data collection is also crucial to successful and smooth project management. As the amount of collected data grows, processing the data that is increasingly massive, multi-sourced, and heterogeneous in nature becomes a challenging task. In the field of the built environment, most of the raw data are unstructured and hence not directly usable. Relying on traditional methods like rule-based systems and expert intervention to extract valuable information from massive data and convert it into structured data requires a lot of time and labor, which cannot meet the ever-growing business needs.

In view of the exponential data growth, knowledge mapping tools have been developed to aid knowledge management and decision-making throughout the project management to boost efficiency and effectiveness in communications. Developing big data and knowledge management tools will be beneficial to facilitate scholars, practitioners, and decision-makers to quickly acquire knowledge from existing projects and make reasonable decisions in the field of the built environment.

One way to manage knowledge intuitively is by using Knowledge Graph (KG), which is proposed to organize and present structured knowledge. It links entities together and defines relationships between entities explicitly to facilitate understanding and knowledge management. For instance, there are many jargons in the built environment field. Extracting these professional terms as entities and the corresponding relation types manually can achieve an accurate yet time-consuming result. Considering scalability would be a major issue in the future, an automated KG construction is preferred to tackle the scalability problem, as well as maintain an accurate entity recognition and relation classification bootstrapped by expert annotations.

As a way to represent structured knowledge, KG was first proposed by Google and delivered as a product in 2012 to improve their search engine and enhance users' search experience. Since then, it has been widely adopted in various information systems for efficient knowledge management. With the rapid research and development in KG in recent years, it can be categorized into two types: general KG and domain-specific KG.

The general KG often contains knowledge across different sectors and scenarios, most of which are common knowledge. A general KG is therefore suitable for recommendation system and Question Answering (QA) system for general users. Existing general KGs such as Freebase (Bollacker et al., 2008), DBpedia (Auer et al., 2007), YAGO (Suchanek et al., 2007), ConceptNet (Liu and Singh, 2004), and Microsoft Concept Graph (Ji et al., 2019) have shown strong potential in improving knowledge acquisition efficiency in many applications.

Compared with the general KG, a domain-specific KG presents specialized knowledge for a specific industry. The target users are experts instead of the public. Hence, it sets a strict requirement in delivering accurate domain-specific knowledge to assist various complex analysis applications and decision support. Entities and relations inside a KG must contain more precise and professional knowledge. At present, domain-specific KG has been widely built in many fields, including medical domain (Ernst et al., 2014; Shi et al., 2017; Goodwin et al., 2013; Rotmensch et al., 2017), cyber security domain (Jia et al., 2018), financial domain (Liu et al., 2019; Elnagdy et al., 2016), common sense domain (Zhang et al., 2020; Fang et al., 2021) and news domain (Rospocher et al., 2016; Rudnik et al., 2019; Ciampaglia et al., 2015). These domain-specific KGs could help users gain specialized knowledge.

KGs have also been applied to different domains with real-world applications. Researchers from the cybersecurity field have used the cybersecurity KGs for cyber defence (Piplai et al., 2020) as an example. A formal geographic knowledge representation model was proposed in the geography sector to deepen the mining of correlated information corresponding to geographic features (Wang et al., 2019). When it comes to the higher education sector, a visualization of the curriculum-based knowledge mappings was used to cope with the increasingly complex teaching system (Yu et al., 2021). There are also feasibility studies on industrial maintenance applications supported by KG to investigate machinery failures and recommend repairing approaches for shield tunneling machines (Qin and Jin, 2020). Besides, KGs are also applied in the oil and gas industry to integrate domain-specific knowledge and manage natural resource exploration, thus improving exploration and development management in the industry (Zhou et al., 2020). As for the public service, Filtz and Kirrane (2021) proposed a method to transform the Australian legal information system into a legal KG that contained various legal documents in a machine-readable format. Moreover, KGs have also shown great potential in improving the efficiency of projects in the built environment, such as a KG proposed by Pauwels (2014) offering higher interoperability for supporting decision-making during the building life cycle. Olawumi et al. (2019) developed a Building Information Modelling (BIM) project information management framework closed to the KG to manage the information of the construction projects effectively.

In this paper, various methods have been adopted to construct a KG for the built environment field. The aim is to build a KG with high accuracy that allows scholars, practitioners, and decision-makers to acquire knowledge easier. The major achievements attained include utilizing the large-scale text data to build a domain-specific KG, which helps reduce labor cost and time spent in the creation, in addition to having achieved promising results. Compared with traditional data management methods in the built environment field, the KG approach is more intuitive and has better scalability for knowledge acquisition and management.

In the section below, the paper starts by introducing related works done in recent KG development and a brief overview of the structure of a general KG. Next, a KG in the built environment field was planned to develop. Journal paper abstracts in the built environment field were used as a corpus for developing a KG because it is an excellent way to capture highlights and insights of a paper. By using the proposed KG development methods, plenty of entities and relations could be extracted from the abstracts to construct the KG. Then, a visualization system to demonstrate the KG was established. As the KG contained large-scale information, a visualization system was necessary to be developed for users to interact and search efficiently. Last but not least, the accuracy of the developed KG and the training model was validated by comparing the results from system prediction with manual annotations. The limitations and future research directions were presented at the end of the paper.

## 2 Related works

The KG construction process could generally be divided into two core steps, data collection and information extraction. The data collection step helped filter out undesirable data and retain valuable data, while the information extraction step revealed meaningful data and transformed it into a structured data format.

### *2.1 Dataset collection*

During the dataset collection stage, the main challenges lay in the unknown relation types to be defined upon data collection. Unsupervised machine learning algorithms such as clustering could detect noticeable patterns among entities and reveal prominent relations in a large-scale corpus. Hasegawa et al. (2004) first proposed the use of an unsupervised machine learning approach for relation extraction based on the assumption that pairs of identical semantic entities had similar contextual contexts. Rozenfeld and Feldman (2006) proposed a method to cluster entity pairs in the same corpus and eliminate candidate pairs with multiple relations to improve Hasegawa's hypothesis, in addition to using a contextual feature-based model to enhance the performance of relation extraction greatly. Plank and Moschitti (2013) used some artificially constructed features such as word clustering and dependency trees to achieve domain adaptation through these domain-independent features.

### *2.2 Information extraction*

For the information extraction step, various existing machine learning methods are available to accomplish the task. Roth and Yih (2004; 2007) formulated the information extraction task as an Integer Linear Programming (ILP) problem to solve information extraction tasks. Besides the linear programming approach, semi-supervised learning is also commonly used to extract useful information from raw data. Training on a small number of seed-labeled training set samples iteratively can then be used to predict a large number of unlabeled samples repeatedly and refine prediction results whenever needed, significantly reducing manual labeling efforts to obtain a classification model. The Dual Iterative Pattern Relation Extraction (DIPRE) system (Brin, 1999) was the earliest proposed semi-supervised entity relation extraction based on bootstrapping on a few labeled samples. This method used a small number of book entity pairs with the format (author, title) as seeds in the early iterations, then automatically obtained new information from the World Wide Web through continuous iterations. Its techniques for acquiring new information can be adopted in Yang et al. (2021)

Meanwhile, Mintz et al. (2009) instead proposed leveraging distant supervision signals from the remote knowledge base to perform entity relation extraction tasks. By automatically aligning the data with the information in the remote knowledge base, it is then used to label the massive data in the open domain to obtain labelled samples.

The concept of open information extraction was proposed by Etzioni et al. (2008) along with the first domain-independent Open Information Extraction (OIE) system, TextRunner, which could be extended to a large-scale Web corpus. Fader et al. (2011) defined two lexical and grammatical constraints on the binary relation expressed by verbs and proved that the problem of incoherence and insufficient information in the Open IE system could be improved by adding these constraints.

In recent years, deep learning has been the most popular approach to improve information extraction performance further. The entity relation extraction method based on the Recursive Neural Network (RNN) was first proposed by Socher et al. (2012) and achieved satisfactory performance. Zeng et al. (2014) demonstrated that using the Convolutional Neural Network (CNN) for relation extraction can remedy complex data pre-processing and feature engineering, requiring only word vectors as the initial input. Zeng et al. (2015) also used a multi-instance learning method to alleviate the noise problem in distant supervision with a CNN model. Jiang et al. (2016) proposed a multi-instance, multi-label convolutional neural network model to relax the at-least-once hypothesis and formulate the relation extraction task as a multi-label problem, which solved the relation overlapping problem. Miwa and Bansal (2016) first used neural network methods to represent entities and relations jointly. Katiyar and Cardie (2017) improved the previous work by leveraging the attention mechanism with a two-way LSTM to extract entities and classification relations together. Despite the promising prediction performance brought by deep learning approaches, these methods were based on supervised learning, which requires a lot of labelled data in advance. Few-shot learning techniques can help maintain accuracy yet with less labelled data. Gao et al and Han et al. (2018) proposed a dataset containing 70,000 instances on 100 relations and transferred few-shot learning into relation classification. Kolluru et al. (2020) presented a neural open information extraction (OpenIE) system with an iterative label method that balanced the computation cost and extraction quality. Huguet Cabot and Navigli (2021) proposed an autoregressive seq2seq model called REBEL to perform the end-to-end relation classification by taking the triples as a sequence of text.

## 2.3 *Knowledge graph used in the built environment*

As KG can intuitively organize entities and relations, it is suitable for domain knowledge management and visualization. Pauwels and Costin (2022) analyzed the availability of KG representation for data in the built environment, including the product data, sensor data, and 3D geometric data. Their findings indicated that KG is useful for product data management and visualization. As for the sensor data, except for direct observations from sensors, their context to the built asset is suitable for KG representation too. When it comes to the 3D geometric data, representing Constructive solid geometry (CSG) in KG however made little sense, but a reference to a condensed mesh representation was a good alternative.

With the recent development in KG, many ontologies were built for the Architecture, Engineering, and Construction (AEC), smart building domains, etc., with a different focus on the categories of data in the built environment field. In the AEC domain, the IFC (Industry Foundation Class) Ontology (Pauwels et al., 2017) and Building Topology Ontology (Rasmussen et al., 2021) were built. As for the smart building domain, Brick[1], Haystack Tagging Ontology (Charpenay et al., 2015), SAREF4Building Ontology (Poveda-Villallon and Garcia-Castro, 2020), and Real Estate Core ontology[2] were also set up. Yet, there is still a lack of a whole concept ontology and KG in the built environment to facilitate scholars, practitioners, and decision-makers to manage knowledge systematically.

---

[1] Brick. A Uniform Metadata Schema for Buildings. https://brickschema.org/

[2] https://www.realestatecore.io/

# 3 Research Methodology

KG is a knowledge base that presents data as a graph to describe entities and their relations. It adopts an "entity-relation-entity" triple format as the primary component unit. Entities are linked to each other through relations to form a graph that describes the complex concepts of things in the real world. In other words, a KG is a graph made of explicitly defined nodes and edges. To illustrate the idea of a KG, a typical subgraph is shown in Figure 1. Entities are presented as nodes, representing concepts and things in the real world, whilst edges are relations, indicating the relations between entities and arrows indicate the relation's direction. Taking 'thermal comfort' as an example, it was affected by the 'acoustic conditions' and could be analyzed by 'physical measurements'. Such knowledge can be represented by the triple 'acoustic conditions, affect, thermal comfort' and other existing triples such as 'physical measurements, analyze, thermal comfort'.

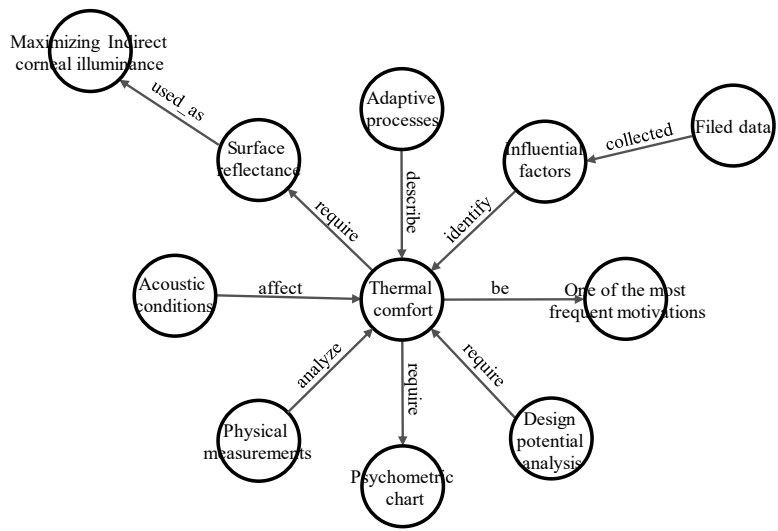

**Figure 1.** An example of a typical KG subgraph.

The overall research methodology flow chart is shown in Figure 2. The first step was data acquisition and processing. Around 85,000 abstracts of paper in 54 journals in the built environment field were acquired in this step. Then, 10,000 pieces of abstract were sampled and pre-processed for relation analysis to determine the schema for the Built Environment Knowledge Graph (BEKG) to obtain a more accurate and comprehensive KG. 50 instances of each relation in the schema were then annotated, followed by obtaining a dataset that can be used for Named Entity Recognition (NER) and Relation Classification model training. Upon deploying the trained model on the unlabelled data, the information of the remaining journal abstracts was extracted. Finally, all extracted entities and relations were used to develop the BEKG.

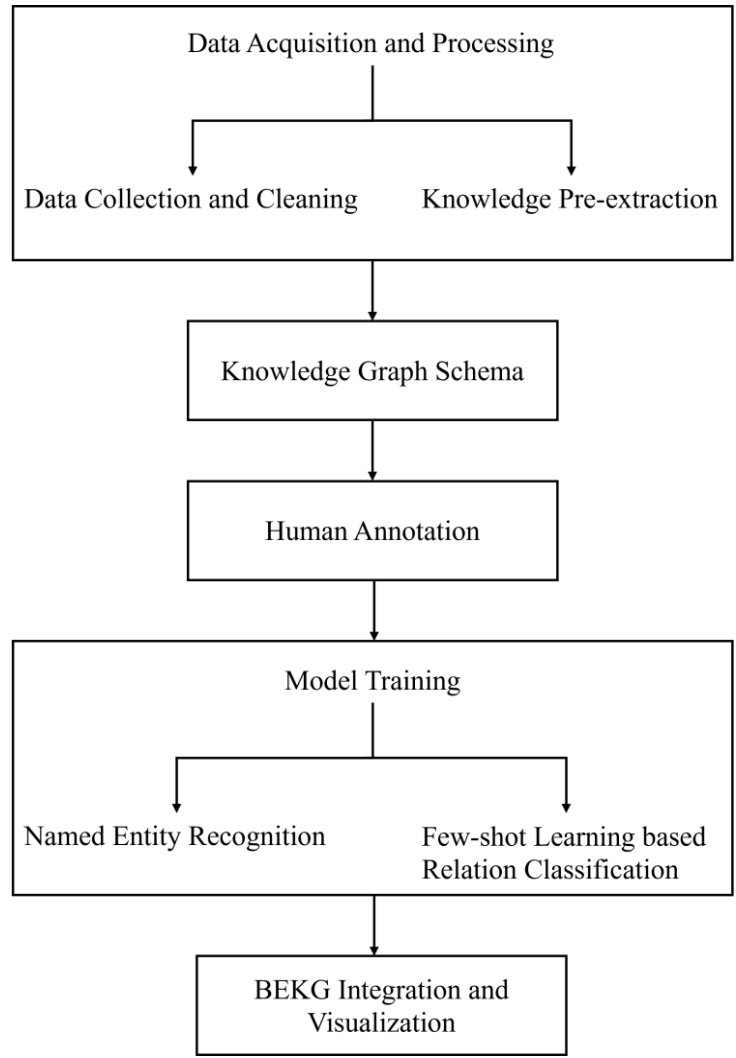

Figure 2. Research methodology flowchart.

*3.1 Data Acquisition and Processing*

Journal abstracts in the built environment field were the base of the BEKG construction presented in this paper. These raw data however required data preprocessing before putting data into knowledge graph construction, the overall steps include data collection, cleaning, and pre-extraction.

*3.1.1 Data Collection and Cleaning*

Publications in the built environment field revealed important insights into the research work in this field. Abstracts were important parts of journal papers, as they summarized research elements, including research problems, methodologies, and results. In addition, abstract data can be easily collected due to the growth of electronic publication. Therefore, abstract data were chosen as the corpus of the BEKG development. The abstracts of journal papers in the built environment field were collected via the Microsoft Academic Graph (MAG) database (Wang et al., 2020). The MAG contained 166,192,182 papers with rich properties, such as the abstract, authors, title, journal, etc.

About 85,000 abstracts in 54 reputable journals in the built environment field were gained from the Microsoft Academic Graph's Azure API. These were top journals selected from the Excellence in Research Australia (ERA) 2018 Journal list provided by the Australian Research Council. All selected journals belong to the Field of Research (FoR) of 1202 (Building) or 1205 (Urban and Regional Planning).

Table 1 shows the Top-5 journals with their amounts and FoR. The earliest abstract data was published around 40 years ago. Except for the data above, the titles and authors were also retrieved from the MAG. The abstract data collected from MAG were in a compressed format. Therefore, original abstract data were obtained after the decompression. Some data cleaning methods were also adopted to remove the escape characters and redundant punctuation marks or characters to ensure the quality of abstract data.

**Table 1.** Top-5 Journals with the quantity of accessed abstract data.

| Journal name | FoR | No. of abstracts collected | Earliest year published in | Latest year published in |
|---|---|---|---|---|
| Energy and Buildings | 1202 | 9,759 | 1977 | 2021 |
| Building and Environment | 1202 | 7,612 | 1976 | 2021 |
| Cement and Concrete Research | 1202 | 7,553 | 1966 | 2021 |
| Environment and Planning A | 1205 | 5,336 | 1969 | 2020 |
| Land Use Policy | 1202 | 5,042 | 1984 | 2021 |

*3.1.2 Knowledge Pre-extraction*

After obtaining the clean abstract data, pre-extraction was needed to bridge the gap between these data and the dataset for model training. This is to create a candidate set containing lots of pre-extraction triples to improve the human annotation efficiency and provide a corpus to construct the KG schema. During the pre-extraction, Ollie (Schmitz et al., 2012) was selected for its efficiency. It can automatically identify and extract the entities and relations from English sentences.

For the first step, 10,000 cleaned abstracts were randomly sampled from the whole data introduced in the last section to ensure the KG schema's generalization. Then, each sentence of the abstract was sent to Ollie to retrieve a large number of triples by extracting relations mediated by nouns, adjectives, and more from its learned pattern templates. Then, a context-analysis module was conducted to add the contextual information from the sentence to these triples. Finally, a confidence score for each triple was given by a confidence function.

As Figure 3 shows, the entity in blue and red indicates the head and tail entity respectively. The yellow box represents the predicted relation, and the black box indicates the confidence score of the extraction result. To ensure the quality of the triples, the sentence output which gives the highest confidence was adopted. Finally, 60,318 pre-extracted triples (head, predicate, tail) were obtained from the 10,000 abstracts, and these triples were used for further processing to determine the BEKG schema.

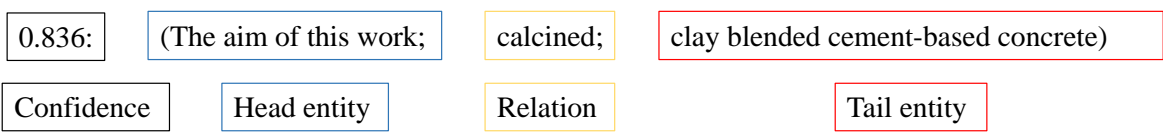

**Figure 3.** One Ollie pre-extraction result of a sentence.

## 3.2 Knowledge Graph Schema

During the pre-extraction step, lots of triples were extracted from input sentences. Yet, these extracted entities and relations were too redundant to be directly used for the model training and implementation. Therefore, the KG schema played a key role in solving this problem. Every KG has its own schema, as it defines the data and relation type in it. In general, a KG schema, which is also an ontology layer, has four definitions: 1) The knowledge in the schema is universally recognized; 2) The description of things is conceptual and has precise specifications; 3) The knowledge in the schema can be formalized.

Following the definition above, the BEKG schema was expected to reach the following goal. The relations defined in the schema should be relevant to the built environment and comprehensive enough to cover as much knowledge as possible. At the same time, the relations defined in the schema should be extensive enough to alleviate the redundancy, which implies that lots of relations of Ollie pre-extraction data in similar semantics should be considered as the same type. Last but not least, it was important to ensure the precision of the schema for the better performance of the trained model and final extraction.

Primitively, an attempt to construct the schema by statistics analysis manually was made. The top 30 frequent types of relation were selected from the Ollie pre-extraction results. However, these types of relations were not extensive and precise enough to reach the goals above.

Therefore, the NLP tool (Reimers and Gurevych, 2019) and the clustering algorithm were adopted. The whole process of developing the schema is shown in Figure 4. The NLP tool was for calculating the text embeddings, transforming the texts from a high-dimensional space into low-dimensional vectors for clustering calculation. Upon obtaining the embedding of the entities in Ollie triples results, the clustering algorithm was adopted to gather the different entities with similar semantics into a cluster. Then the entities in the original Ollie triples were updated depending on the clustering results. Finally, the updated Ollie triples were transformed into embeddings and clustered again to acquire the knowledge graph schema.

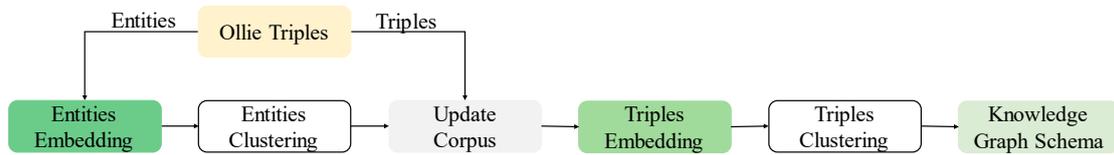

**Figure 4**. The process of building a KG schema.

The chosen NLP tool for calculating the text's embedding was a model with the best performance in Sentence-BERT (Reimers and Gurevych, 2019), called 'BERT-STSb-large', which adds a mean pooling operation to the BERT-large model output to obtain a fixed-size sentence embedding. It was trained with the max sequence length of 128 on the STS benchmark (Cer et al., 2017), a popular dataset to evaluate supervised Semantic Textual Similarity (STS) systems. By feeding the triples into the model, it outputs an embedding with a dimension of 768, which significantly improved the efficiency but maintained the effectiveness.

As for the clustering algorithm, the k-means (Lloyd, 1982) is adopted. The optimization criterion of the k-means algorithm is expressed as Eq. (1):

$$J = min \sum_{i=1}^{k} \sum_{x \in C_i} ||x - u_i||^2, u_i = \frac{1}{|C_i|} \sum_{x \in C_i} x, \tag{1}$$

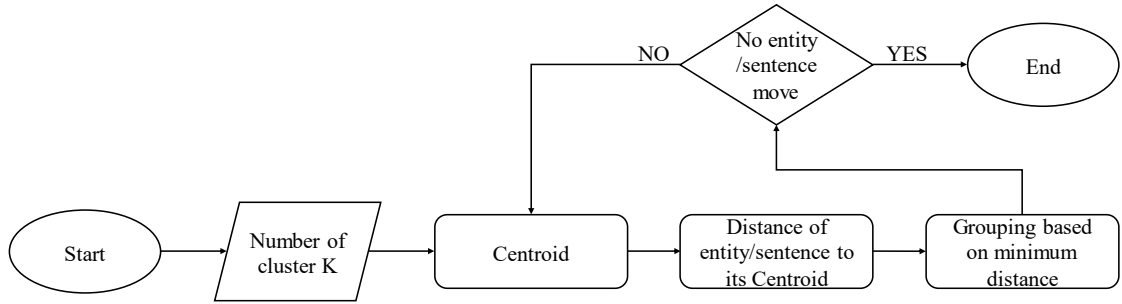

**Figure 5.** The flow chart of the clustering algorithm.

where $x$ denotes the embeddings of the entity or triple, $C = \{C_1, C_2, \ldots, C_k\}$ denotes the set of clusters. $u_i$ denotes the centroid of each cluster $C_i$. $J$ was optimized iteratively to reach a minimum until $u_i$ was fixed so that the similarity of samples in the same cluster becomes as high as possible. The flow chart of the clustering algorithm is shown in Figure 5.

**Table 2.** Examples of some types of relations and their instances, blue for the head entity and red for the tail entity.

| Relation type | Top-3 Similar Head Entities | Top-3 Similar Tail Entities |
|---|---|---|
| **Relation_1**: Be Made | **Entity_198**: aluminum metal<br><br>**Entity_210**: cement<br><br>**Entity_450**: Portland limestone | **Entity_134**: market expansion process<br><br>**Entity_138**: clinker<br><br>**Entity_264**: Climate condition moisture source |
| | **Triple:** Cements<**Entity_210**> made from<**Relation_1**> clinker<**Entity_138**>. | |
| **Relation_2**: Lead to | **Entity_126**: pore water pressure transducer<br><br>**Entity_104**: steam curing<br><br>**Entity_308**: paper analysis cost | **Entity_283**: work experience performance.<br><br>**Entity_9**: lower porosity and fewer micropores.<br><br>**Entity_61**: activity engagement. |
| | **Triple:** Steam curing<**Entity_104**> led to<**Relation_2**> lower porosity and fewer micropores <**Entity_9**>. | |
| **Relation_3**: Related to | **Entity_297:** transformative adaptation<br><br>**Entity_402:** discomfort probability<br><br>**Entity_46:** transitions theories | **Entity_266:** wall friction effect<br><br>**Entity_311:** manufacturing industry<br><br>**Entity_81:** regional scholarship and practice |
| | **Triple:** transitions theories<**Entity_46**> related to<**Relation_2**> regional scholarship and practice<**Entity_81**>. | |
| **Relation_4**: Improve | **Entity_308:** paper analysis cost<br><br>**Entity_250:** study contribution<br><br>**Entity_371:** The addition of fly ash | **Entity_130:** survey inquiry research<br><br>**Entity_386:** model predictive controller<br><br>**Entity_146:** the microstructure of MPC |
| | **Triple:** The addition of fly ash<**Entity_371**> improves<**Relation_4**> the microstructure of MPC<**Entity_146**>. | |
| **Relation_5**: Include | **Entity_254:** accuracy measure exposure<br><br>**Entity_349:** field data information<br><br>**Entity_29:** Modern concrete construction | **Entity_166:** heat strain prediction<br><br>**Entity_419:** rate estimation method<br><br>**Entity_173:** the use of a variety of special concretes |
| | **Triple:** Modern concrete construction<**Entity_29**> includes<**Relation_5**> the use of a variety of special concretes<**Entity_173**>. | |

The number of entity clusters and triple clusters to be used follows the approach in Wang et al. (2018), which was determined by dividing the total number of entity clusters and triple clusters by 5 respectively. Before triples clustering, the entity closest to the centroid was chosen to represent each cluster respectively, then all entities in the corpus were updated by their representative entity. The relations in the final knowledge graph schema were obtained similarly.

Throughout the implementation, duplicated entities in about 58,000 Ollie pre-extracted triples were removed firstly to obtain 9,800 entities which were then clustered into 56 clusters. The Ollie pre-extracted triples were updated into approximately 3,000 triples according to the entities clustering results. Finally, 29 relation types were gained to construct the KG schema after the triples clustering. Table 2 shows some relation types in the schema that were presented. The entities in the row 'Top-3 Similar Entities' are the top-3 similar ones with the representative of each entity cluster. Each example in the line 'Triple' denotes one triple in the triple cluster, updated depending on the entities clustering results.

*3.3 Human Annotation*

The next step was annotating the data following the obtained KG schema to build a dataset. To obtain higher quality triples to create a dataset for model training, several annotators were organized to annotate the triples on the Brat platform (Stenetorp et al., 2012), a web-based annotation tool for employing each type of annotation on text easily and effectively. The Brat was used to structure the original unstructured text to obtain the annotation corpus required for the named entity recognition model and relation classification model training.

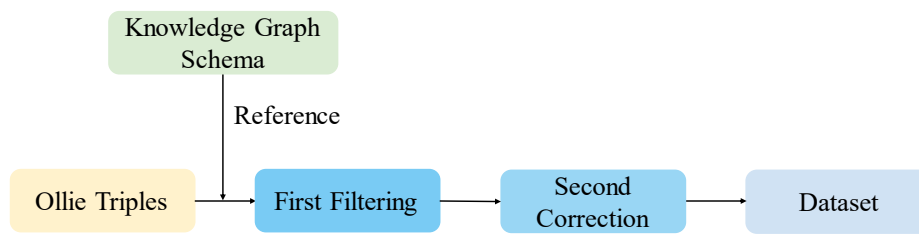

**Figure 6.** Workflow of human annotation.

The human annotation workflow is shown in Figure 6. First, depending on the BEKG schema, Ollie triples from pre-extraction were classified to each relation. Then, several well-educated annotators were organized to filter the Ollie extraction results of high-quality belonging to each relation using the Brat. The Brat illustrated each Ollie triple by displaying the original sentence with two highlighted entities linked by a line with their relation above.

During the first filtering, annotators aimed at looking for the good Ollie results whose entities and results corresponded to the built environment without mistakes. However, the amount of these well-annotated Ollie results was small. For other Ollie extraction results, the annotator had to correct some entity annotation mistakes that appeared in them.

In general, annotators were asked to ensure the high-quality instance with precisely annotated entities and the correctness of relations labelled. As the whole process relied on subjective recognition, some criteria were provided for them during annotating.

Specifically, the clause and adverb in the annotated entities were removed to lower the redundancy of the annotation. In addition, the coordinate entities in one annotated entity were split, which would not confuse the NER model during training. A few examples of the high-quality Ollie extraction results obtained after the first filtering are shown in Figure 7.

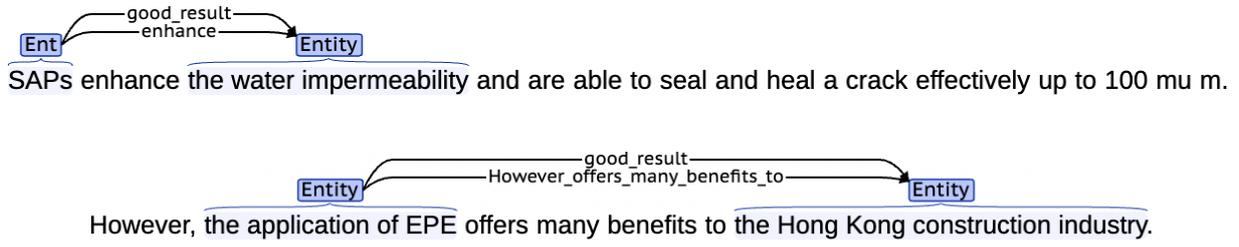

**Figure 7.** Examples of human annotations.

After the first filtering, an annotator who was not involved in the first round of filtering would be asked to validate annotations and correct them if necessary to ensure mutual agreements among annotators. The annotation costs time and manpower heavily. However, it was difficult to find annotators familiar with the BE domain. Therefore, considering the balance of efficiency and effectiveness, the number of well-annotated instances for each relation was set as 50.

Finally, 1,450 high-quality instances in 29 relations were obtained for developing relation classification datasets. Regarding the dataset split method adopted in Han et al. (2018), 29 relations were shuffled first, then 18, 5, and 6 of them were allocated to relation classification training, validation, and testing set correspondingly.

As the performance of the NER model plays a crucial role in the final extraction results, 550 more annotated instances were added to the NER dataset. 1,600 and 400 instances were randomly selected and allocated to the training and validation set respectively. There was no test set in the NER dataset as the model performance was evaluated by humans at the extraction phase, which could increase the amount of data in the training set. The statistics of these two datasets are shown in Table 3.

**Table 3.** Overall statistics of the annotated dataset.

|  | Relation Classification Dataset | NER Dataset |
| --- | --- | --- |
| Training Set | 900 | 1,600 |
| Validation Set | 250 | 400 |
| Test Set | 300 | - |
| Overall Instances | 1,450 | 2,000 |
| Relations Type | 29 | - |
| Instances of each relation | 50 | - |

*3.4 Model Training*

Training models using the annotated dataset were necessary to achieve better performance. According to the BEKG schema, two datasets containing more than 1400 instances were built. There was a pair of entities and their relation in each instance. Besides, a suitable named entity recognition model and relation classification model were in need to extract entities and relations in the data extraction stage.

*3.4.1 Named Entity Recognition*

Named entity recognition refers to the task of recognizing entities in an input sentence. The named entity recognition model could annotate each token in the sentence using one of the 'BIO' sets. 'B' means the beginning token of the entity. 'I' represents the token inside the entity, and 'O' means the last token in the entity or token that does not belong to any entity.

The named entity recognition was performed by BERT-CRF, a combination of BERT (Devlin et al., 2018), which is a large masked language pre-trained model for text representation, and CRF (Lafferty et al., 2001), an objective of structured output prediction. The forward of BERT-CRF could be expressed as Eq. (2):

$$\text{Pred} = C_{\text{decode}}\left(l\left(d(B(X))\right)\right), \tag{2}$$

where $B$ denotes the BERT model, $d$ and $l$ denote the dropout and linear layer respectively. The linear layer outputs the emission score of each token in the input sequence for the 'BIO' label. $C_{decode}$ denotes the Viterbi algorithm for finding the best tag sequence given an emission score tensor.

The BERT-CRF was trained to maximize the log-probability of the correct tag sequence. Therefore, the loss function of the model was designed as Eq. (3):

$$\ell = -\text{C}\left(\log(S(emission))\right), \tag{3}$$

where $emission$ denotes the output of the linear layer, $S$ denotes the softmax layer, and $\text{C}$ denotes the CRF layer.

A demonstration of how the named entity recognition model worked is shown in Figure 8. BERT-CRF took the tokenized sentence as input and output each token's 'BIO' annotation. A model was trained on the dataset since the pre-trained model's performance was not satisfied. The dataset contained 2,000 instances, each consisting of a sentence and two annotated pairwise entities.

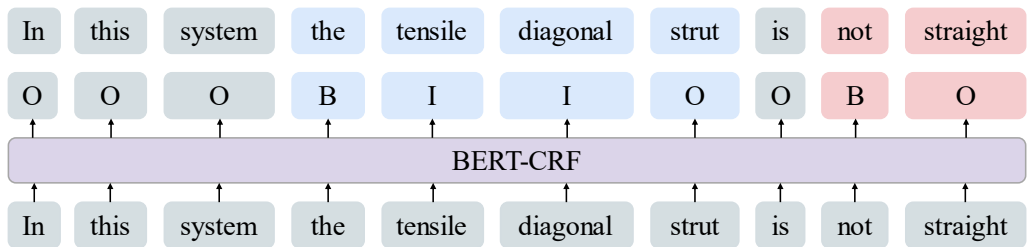

**Figure 8.** The visualization of named entity recognition: blue for the head entity and red for the tail entity.

*3.4.2 Few-shot Learning-based Relation Classification*

Compared with the named entity recognition task, the relation extraction task requires a model with a better generalization. The few-shot learning method aims to solve a problem through a few annotated samples, which meets the requirement for the model to perform a great generalization with a limited amount of data. Therefore, BERT-Pair (Gao et al., 2019) was taken as the few-shot learning-based model for the relation extraction task. It was modified based on the BERT to consider the relation extraction task as a few-shot relation classification task. During the few-shot learning, a support set and a few query sets are needed. The 'N-way K-shot Q-query' setting is usually adopted for constructing support and query sets. These parameters indicate that N out of all classes in the dataset would be selected. Then, K instances of each class chosen are obtained for constructing a support set $S$. It can be expressed as Eq. (4).

$$S = \begin{cases} (x_1^1, r_1), (x_1^2, r_1), \dots, (x_1^K, r_1), \\ \dots, \\ (x_N^1, r_N), (x_N^2, r_N), \dots, (x_N^K, r_N) \end{cases} \quad (4)$$

where $r_N$ denotes the Nth class and $x_N^K$ denotes the K-th instance of the N-th class in the support set. Then, Q instances were picked from each selected class's training instances, excluding instances that have been selected for the support set to construct N query sets. It can be expressed as Eq. (5).

$$Q = \begin{cases} (y_1^1, r_1), (y_1^2, r_1), \dots, (y_1^Q, r_1), \\ \dots, \\ (y_N^1, r_N), (y_N^2, r_N), \dots, (y_N^Q, r_N) \end{cases} \quad (5)$$

where $y$ denotes the instance different from instances in the support set. $y_N^Q$ denotes the Q-th instance of the N-th class in the support set. To illustrate these two sets, a demonstration of constructing a "5 way 1 shot 1 query" setting is shown in Figure 9.

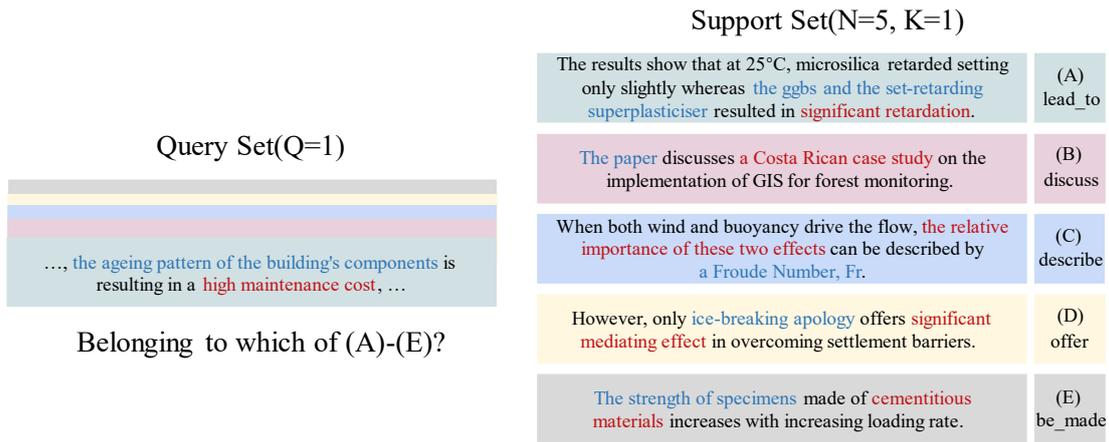

**Figure 9.** A demonstration of the '5-way 1-shot 1-query' setting for a support set and query set during training: blue for the head entity and red for the tail entity.

To obtain the support set and query set, each query and support instance were first tokenized, then the embedding was calculated for each token. The embeddings of each tokenized query instance were concatenated with each support instance's tokenized embedding, as shown in Figure 10. BERT-Pair model based on the BERT sequence classification model is used to calculate the relation similarity between two instances in each concatenated embedding whose max length is 128. The whole process of the similarity calculation during each query can be expressed as Eq. (6).

$$O_{pred} = \max_{r \in \{r_1,...,r_N\}} \frac{1}{K} \sum_{j=1}^{K} (BP(y_r, x_r^j)) \qquad (6)$$

where $O_{pred}$ denotes the score of the predicted label and $BP$ denotes the BERT-Pair model. For better understanding, Figure 10 demonstrates the calculation process of a 5-way 1-shot setting. Each row in the Figure 10 table indicates the similarity score of a query instance with each class in the support set. The highest score for the first row was 8.6. Therefore, the first query is classified into the 'lead_to' class.

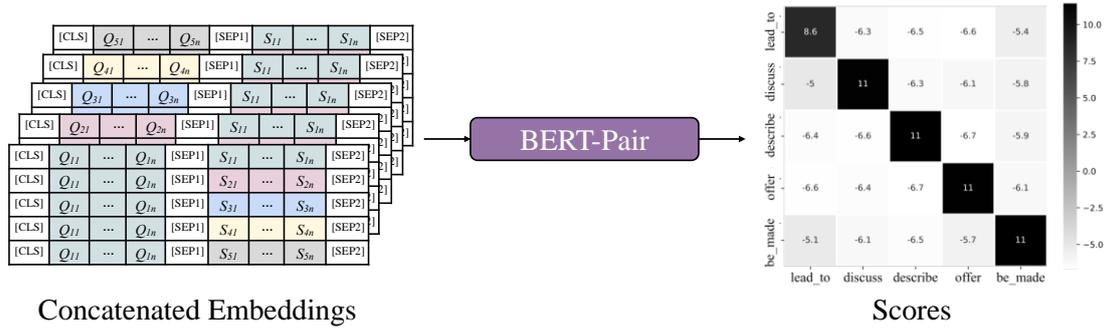

**Figure 10.** The visualization of relation classification using the BERT-Pair model.

At the stage of model training, both the BERT-CRF and BERT-Pair models were trained on Titan Xp. During the BERT-CRF training, the batch size was set to 16. The learning rate started from 5e-8 and the Adam optimizer (Kingma and Ba, 2014) was adopted for model optimization.

For the BERT-Pair training, the same N-way K-shot settings were kept throughout the training and evaluation. Due to the limited number of relations in our constructed dataset, 5-way 1-shot was the optimum choice from the recommended setting in Han et al. (2018) when evaluating the model performance on the validation set with only 5 relations. As a result, the 5-way 1-shot learning strategy was adopted for BERT-Pair model training. Then, 2e-5 was selected as the initial learning rate. Warmup strategy was not adopted during the model training as both the BERT-CRF and BERT-Pair were finetuned using the pre-trained BERT models. The batch size was 2 due to GPU memory limitations.

*3.5 BEKG Integration and Visualization*

The methods above are proved suitable to accomplish the goal of developing a BEKG. With the trained model, it can be used to extract the entities and relations in the abstract data. The process of constructing the BEKG can be separated into four parts: entity extraction, relation extraction, post-processing, and visualization.

As the relation classification model requires a pairwise entity as the input, it is necessary to conduct entity extraction prior to the relation classification. The whole process of extraction is shown in Figure 11. During the entity extraction process, all sentences in the journal paper abstracts were first tokenized into the independent token. Then each token in each sentence was passed to the BERT-CRF model, outputting each token's 'BIO' annotation. Since more than two entities may be extracted from a sentence, all the possible entity pairs were extracted and passed into the relation classification model to confirm whether any relation existed before further categorizing into more fine-grained relations. Therefore, these entities should be paired into the pairwise entity as the input of the relation classification model.

After combining the sentence along with the pairwise entity, these inputs were passed into the model to extract the relation between the pairwise entity. During the relation extraction stage, the whole process was regarded as a '29-way 1-shot 1-query' setting few-shot classification. The input sentence and pairwise entities were taken as the query instance. Then, the BERT-Pair model calculated the similarity between the query instance and every 29 support instances and output the most similar relation from the 29 types of relations.

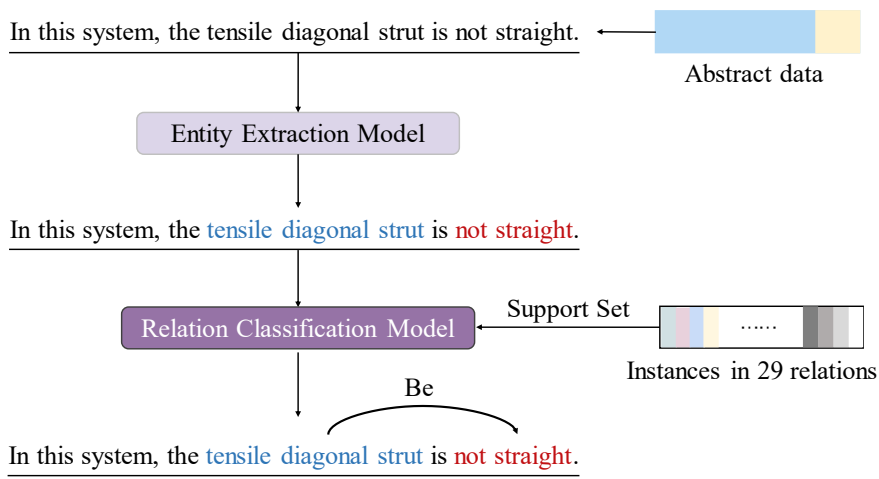

**Figure 11.** An example of the whole extraction process from a sentence during final data extraction: blue for the head entity and red for the tail entity.

Despite having obtained a large number of extracted entities and relations, it was difficult for users to analyze such a diverse and massive amount of data effectively. A practical method is to use the current widely used KG web visualization technique, which could effectively represent information. Figure 12 shows the initial user interface of the visualization system, which combines Navigation Bar, Search Box, and Graphical User Interface. The interface is designed to be user-friendly such that users can search and browse the target graph within a few simple steps.

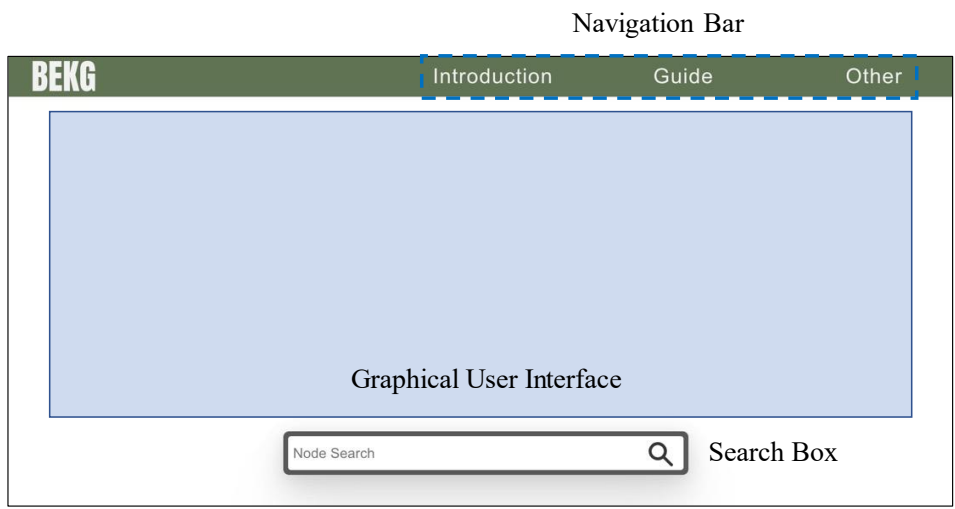

**Figure 12.** The user interface of the KG visualization system.

Specifically, users can switch pages between the Navigation Bar and enter words with interests related to the built environment in the search box. The user interface of the search results is shown in Figure 13. The nodes selected by users from the generated candidates related to the search words will appear in the Graphical User Interface.

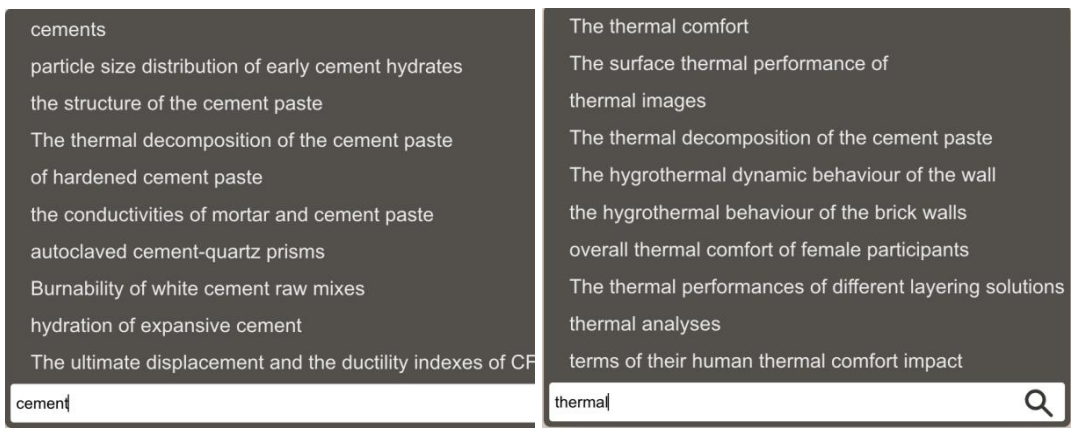

**Figure 13.** Demonstration of the search results gained from the KG visualization system.

Figure 14 shows what the visualization system did when the users interacted with it. After the user clicks the nodes drawn by the front end, it immediately sends the request to the back end. The back end returns the linked nodes and relations back to the front end for visualization. The design of the front end was based on the characteristics of KGs. At the front end, the Vue framework was adopted for the basic graph interface and the D3 framework for the graph layout algorithm. At the back end, all triples were stored inside the Neo4j graph database, assessed by a functional framework called Mybatis. The spring-boot framework provided the basic logistic functions for web applications.

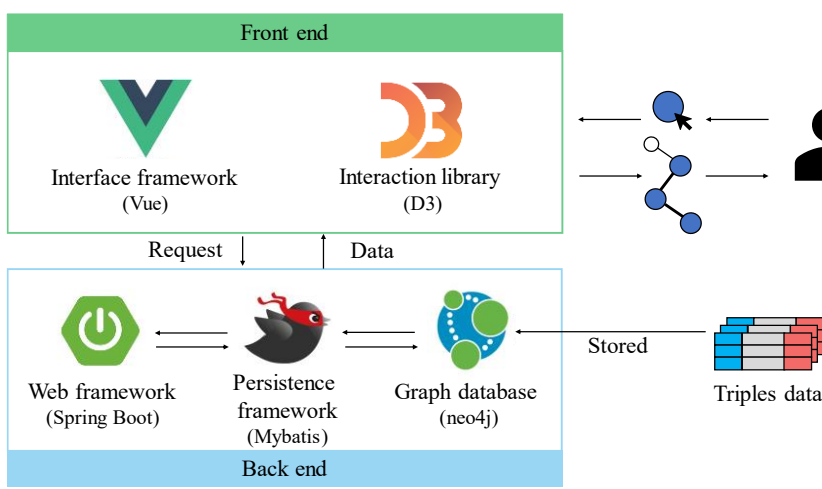

**Figure 14.** Workflow of the KG visualization system.

A sub-graph illustration related to the entity 'cements' of the complete BEKG is shown in Figure 15. The nodes represent entities, and the lines between the nodes represent their relations. At the same time, the source paper title related to each entity can be found in its specific details. In this way, the visualization system offers an efficient way for users to retrieve and trace the research of interest.

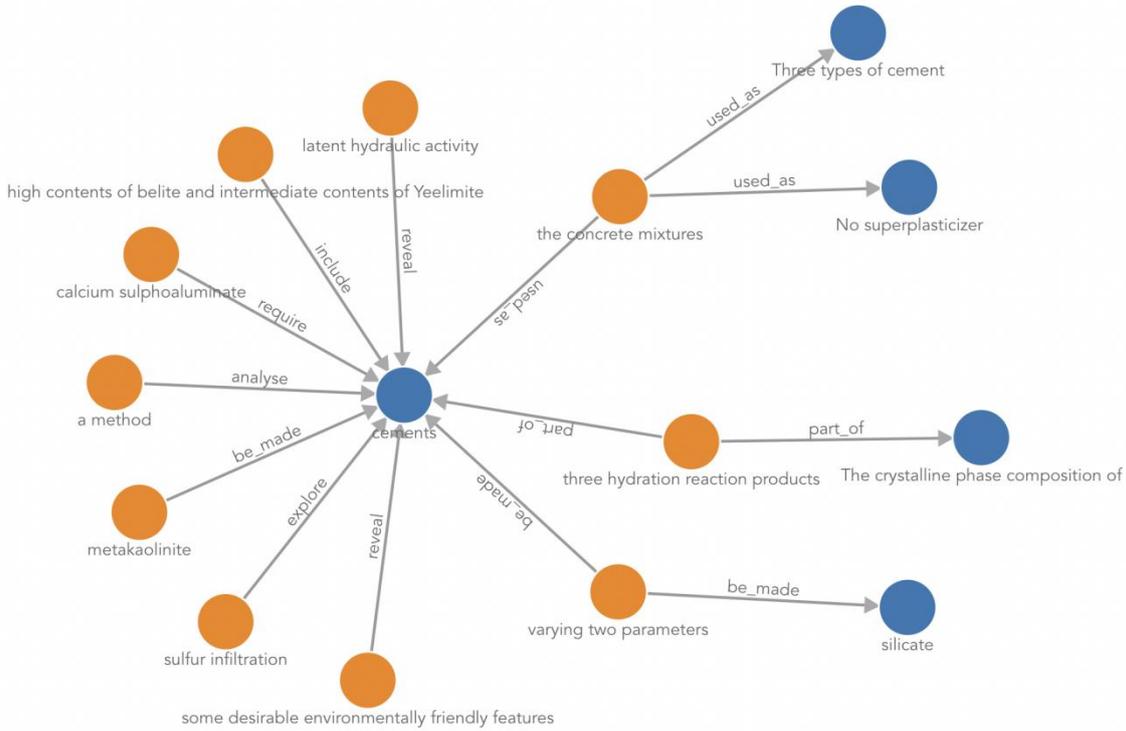

**Figure 15.** Visualizations of Part of the BEKG.

## 4  Result Evaluation and Analysis

The quality of the extracted knowledge was sensitive to the accuracy of each model for extraction. Therefore, having a proper way to evaluate the performance at each stage is vital in the experiments. In the following sections, this paper will analyze the experiment results and elucidate how it improved the model performance.

*4.1 Accuracy of Each Stage*

During the training of the NER model, the validation set containing 400 annotated instances was used to evaluate the model performance. Results are shown in Table 4. Besides the BERT-CRF, other common NER methods were also trained and evaluated to make comparisons with BERT-CRF on the dataset. Four metrics were used to assess the performance of the validation set in different dimensions. The BERT-Span's accuracy was vacant because its output format did not fit the NER problem. The results indicate that the layer of CRF added to BERT as the last layer outperformed other methods on the validation set in nearly all four metrics.

Among these metrics, the Accuracy metric is the ratio between the amount of correctly predicted tokens with all tokens, which implies this metric evaluates the performance on the token level. The Precision metric is an entity-level metric that represents the ratio of the amount of the correctly predicted entities with all predicted entities. The Recall metric is the ratio of the amount of the correctly predicted entities with all labeled entities. The F1-score is a metric taking the harmonic mean of precision and recall.

**Table 4.** Comparison of the accuracy of different NER methods on the validation set.

| Methods | Accuracy | Precision | Recall | F1-Score |
|---|---|---|---|---|
| BERT-CRF | 92.31% | 71.48% | 77.24% | 74.25% |
| BERT-Softmax | 87.49% | 69.15% | 64.36% | 66.67% |
| BERT-Span | - | 72.53% | 62.85% | 67.34% |
| ALBERT-Softmax | 88.23% | 71.22% | 67.63% | 69.38% |
| BERT-BiLSTM-CRF | 91.82% | 68.37% | 73.53% | 70.86% |
| SpanBERT-CRF | 90.79% | 66.31% | 71.48% | 68.80% |

The results in Table 4 do not reveal the model's robustness as the constructed dataset is of a small size. To assess the model performance objectively, a k-fold cross-validation is performed to split the original dataset with 2,000 instances into k folds of the same size to conduct evaluations.

For keeping the same size of the dataset as the original one, the 5-fold setting was chosen which split the dataset into training and validation in the ratio 4:1. Other settings during training and validation were also kept in accordance with previous ones.

The cross-validation results on the validation set are shown in Table 5. It is worth noting that the model performance on fold-2 was worse than on other folds. The standard deviation (STD) of four metrics on the 5-fold cross-validation also showed that the model robustness was affected by the limited size of the dataset slightly. In spite of discrepancies between the averaged results and individual folds in Table 4, the performance was approximately within one standard deviation.

**Table 5.** NER tagging results of BERT-CRF on the validation set using 5-fold cross-validation.

|  | Fold-1 | Fold-2 | Fold-3 | Fold-4 | Fold-5 | Average | STD |
|---|---|---|---|---|---|---|---|
| Acc | 91.62% | 91.26% | 91.82% | 92.14% | 92.24% | 91.82% | 0.40% |
| Precision | 69.03% | 64.55% | 68.97% | 69.33% | 70.78% | 68.53% | 2.34% |
| Recall | 75.70% | 71.12% | 76.37% | 76.40% | 77.55% | 75.43% | 2.50% |
| F1-Score | 72.21% | 67.68% | 72.48% | 72.69% | 74.01% | 71.81% | 2.41% |

As for the relation classification, results were evaluated separately at the training and large-scale extraction stage in two ways. The validation and test set were used to assess the model performance throughout the model training to prevent overfitting. The evaluation results on each set are shown in Table 6.

**Table 6.** Accuracy of relation classification at the training phase.

| Validation Set | Test Set |
|---|---|
| 85.51% | 86.3% |

The 5-fold cross-validation was also conducted on the relation classification task. The division among the training, validation, and test set in each fold reuses the previous fold's allocation with a four-relations right rotation, ensuring each relation was deployed for validation or test set. The allocation of the original dataset was the initial setting for Fold-1.

The results of each fold's validation and test set were obtained after 1,000 iterations of few-shot relation classification. As shown in Table 7, the average accuracy on validation and test set were both close to the accuracy in Table 6.

The standard deviation of accuracy on the 5-fold validation set and test set cross-validation were higher than in the NER task. Some relations containing more types of relation words would be a challenge for the BERT-Pair model to classify which pulled down the accuracy of individual folds. At the same time, as adopting a different way to allocate dataset for each fold following the few-shot learning, a domain gap between the training and validation or test set also affected the accuracy.

**Table 7.** Relation classification results on both validation set and test set using 5-fold cross-validation.

|      | Fold-1 | Fold-2 | Fold-3 | Fold-4 | Fold-5 | Average | STD |
|------|--------|--------|--------|--------|--------|---------|-----|
| Val  | 85.03% | 76.24% | 95.32% | 92.16% | 87.68% | 87.29%  | 7.34% |
| Test | 89.98% | 90.70% | 89.56% | 81.42% | 70.46% | 84.42%  | 8.67% |

At the final data extraction stage, massive instances were obtained. As it was impossible to validate them all manually, 100 instances per relation of the total 29 relations were sampled to be validated instead to alleviate the manual validation workload. During validation, every instance was validated by two different examiners. Examiners gave either 'True' or 'False' to judge the correctness of the relation between two entities in the instance. The validation result was adopted when two examiners reached a consensus. Otherwise, the third annotator was asked to make the final judgment to ensure the correctness and objectivity of the result. The results of this phase are shown in Table 8. It showed that the overall quality of the knowledge in BEKG was acceptable, with accuracy above 80%.

**Table 8.** Accuracy of relation extraction at data integration phase.

|                                              | True  | False | Total | Accuracy |
|----------------------------------------------|-------|-------|-------|----------|
| Mutually agreed                              | 1,965 | 236   | 2,201 | 89.28%   |
| Disagreed and annotated with a third annotator | 456   | 243   | 699   | 65.24%   |
| Total                                        | 2,421 | 479   | 2,900 | 83.48%   |

To further analyze the model's generalization for different relations, human evaluations on each relation were performed with the accuracy metric. Following the statistics result in Table 9, BERT-Pair performed well on most relations yet encountered difficulties on 'part_of' and 'identify'. The 'part_of' relation is comprised of several types of relation words in similar semantics, such as 'composed of', 'formed of', 'served as', and so on. Although 'identify' just had a single relation word, it is a polyseme, resulting in ambiguities in different scenarios and might be hard for the relation classification model to classify. BERT-Pair benefited from the large-scaled pre-trained corpus settings and could however interpret the meaning correctly, covering most of the relations in the dataset while having a slight performance decrease in handling a few relations with rich diversity.

**Table 9.** Accuracy of relation extraction at data integration phase.

| Relation Type | Accuracy | Relation Type | Accuracy |
|---|---|---|---|
| represent | 84.00% | calculate | 71.00% |
| discuss | 73.00% | find | 80.00% |
| draw_on | 91.00% | improve | 91.00% |
| part_of | 66.00% | identify | 68.00% |
| require | 83.00% | be | 90.00% |
| include | 92.00% | explore | 88.00% |
| report | 89.00% | lead_to | 86.00% |
| used_as | 73.00% | reveal | 87.00% |
| affect | 84.00% | related_to | 87.00% |
| consider | 70.00% | produce | 92.00% |
| examine | 84.00% | offer | 88.00% |
| analyse | 74.00% | develop | 86.00% |
| collect | 86.00% | concern | 87.00% |
| be_made | 86.00% | focus_on | 86.00% |
| describe | 92.00% | | |

## 4.2 Overall Statistics of the BEKG

At the final data extraction stage, entity extraction was first applied to each sentence collected to retrieve pairwise entities. The pairwise entities were then passed to the relation classification model along with the sentences that they belonged to, predicting the relation between each pairwise entity. Among the 84,588 pieces of all the abstracts collected, 400,200 sentences were obtained. 771,648 entities were then acquired using the trained BERT-CRF model. These entities were then able to combine 458,726 pairwise entities, each of which was then assembled into an instance with their belonging sentence. Each instance was sent into the BERT-Pair model, which took advantage of the annotated dataset as a supporting set to obtain the final extracted relation. The statistics are shown in Table 10.

**Table 10.** Results statistics at data integration phase.

| Final Data Extraction | Abstract | Sentences | Entities | High-quality Triples |
|---|---|---|---|---|
| Total | 84,588 | 400,200 | 425,650 | 223,443 |

Each triple was attached with a score indicating the possibility of the output. The number of triples harvested over different thresholds was presented as a line graph in Figure 16.

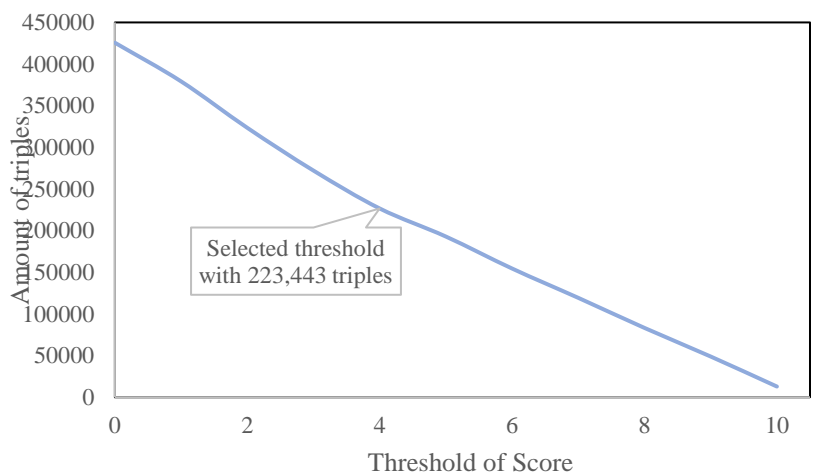

**Figure 16.** Distribution of triples by relation classification score.

**Table 11.** Statistics of triple and entities belonging to each relation.

| Relation Type | Triples | Entities | Relation Type | Triples | Entities |
|---|---|---|---|---|---|
| Include | 29,710 | 40,379 | Related_to | 4,572 | 5,046 |
| Analyze | 3,118 | 4,132 | Draw_on | 2,686 | 2,688 |
| Reveal | 21,620 | 24,490 | Be_made | 3,587 | 4,185 |
| Used_as | 15,030 | 17,789 | Develop | 7,320 | 7,868 |
| Examine | 7,099 | 7,127 | Produce | 3,756 | 4,287 |
| Calculate | 3,089 | 3,938 | Explore | 8,198 | 8,203 |
| Represent | 2,262 | 2,905 | Lead_to | 8,637 | 10,159 |
| Collect | 6,843 | 7,416 | Concern | 1,561 | 1,584 |
| Find | 2,243 | 2,769 | Describe | 4,465 | 4,143 |
| Be | 29,023 | 36,138 | Report | 1,729 | 1,423 |
| Identify | 4,725 | 5,261 | Affect | 6,260 | 7,003 |
| Improve | 2,086 | 2,893 | Part_of | 3,629 | 3,690 |
| Consider | 9,845 | 10,836 | Offer | 3,727 | 3,739 |
| Require | 15,292 | 18,356 | Focus_on | 6,737 | 6,155 |
| Discuss | 4,594 | 3,732 | | | |
| Total | | | | 223,443 | 258,694 |

A graph with 223,443 high-quality triples filtered with an empirical threshold was obtained. The threshold was determined to balance the quantity and quality of the triples extracted. With this threshold, triples with obviously incorrect relations were filtered, which could ensure the preciseness of the BEKG to retain high-quality triples. A breakdown in the number of triples and entities belonging to each relation with the selected threshold was reported in Table 11.

## 5   Conclusion

As lots of knowledge in the built environment field remained siloed in unstructured data format, it gave the scholars, practitioners, and decision-makers a hard time organizing them in a short time. Lacking a proper knowledge visualization and organization could be a barrier to improving the level and efficiency of project management. A KG can transform knowledge in unstructured data into structured data to solve this problem.

In this paper, two well-annotated datasets for named entity recognition (NER) and relation classification (RC) were developed. With the help of a clustering algorithm, two datasets containing 29 types of relations and more than 1,500 instances annotated manually were built. Trained on this dataset, the BERT-CRF and BERT-Pair models were able to attain a test accuracy above 85%. Adopting the trained model to implement final data extraction, the BEKG and its visualization system were presented. This BEKG contained more than 250,000 entities and 220,000 relations, which offered a way to intuitively acquire knowledge in the built environment field when cooperating with the visualization system. In summary, the main contributions consist of the following: 1) an effort for building the graph has been spent to develop a built environment knowledge graph; 2) human annotated datasets for evaluating the NER and the relation extraction components were constructed; 3) a visualization tool to explore the knowledge graph to understand the entities and relations was developed. This is a very practical end-to-end effort in building a domain-specific knowledge graph and could have major impact to the built domain.

The limitation of this research was that the veracity of extraction was unable to reach the human level yet. To illustrate, the relation classification model could not correctly predict all directions of the relations between the pairwise entities, which might confuse users. At the same time, the accuracy of some relations was not ideal enough due to the limited ability of existing models.

This knowledge graph aims to help professionals in this domain accomplish their work more efficiently and effectively. However, the performance of this initially developed KG in real-world applications can be improved with new techniques for solving the existing insufficiency employed.

Specifically, the relation classification model cannot perform well enough on some relations as the model cannot handle the diversity of data in these relations. To solve this problem, enlarging the scale of the accurate data by updating the abstract of the latest paper into the BEKG and designing a more intelligent model are practical ways to improve the model's generalization.

Based on the fundamental research done in this paper, various application scenarios in which the BEKG can be deployed in the built environment areas and for future research directions. For example, according to the research done by Fang et al. (2020), BEKG can be applied to improve workplace health and safety. Also, referring to the research done by Pan et al. (2021), BEKG can be used to improve the construction process management. Finally, the adoption route of an enterprise knowledge graph (Song et al. 2019) is recommended if BEKG is deployed for commercial use.

The BEKG's target audience is the professionals in the built environment areas. Therefore, obtaining more feedback from them is the best way to improve BEKG's user experience in the future. Some professional users will be invited to offer their advice about different aspects of BEKG, including the user interface, the effectiveness of the BEKG on their work, and other requirements that have not been taken into account.

Moreover, the data and code of BEKG have been opened source. Its quality and quantity will greatly improve the research of KG in the BE domain, allowing more researchers and developers in the field to make further contributions.

## 6  Acknowledgment

This paper was supported by the National Natural Science Foundation of China (U20B2053), the Key-Area Research and Development Program of Guangdong Province (2018B010115001), and the Key Laboratory Open Foundation (WDZC20215250118), the GRF (16211520), and the RIF (R6020-19 and R6021-20) from RGC of Hong Kong. We also thank the UGC Research Matching Grants (RMGS20EG01-D, RMGS20CR11, RMGS20CR12, RMGS20EG19, RMGS20EG21).